\documentclass[lettersize,journal]{IEEEtran}
\usepackage{amsmath,amsfonts}
\usepackage{array}
\usepackage[caption=false,font=normalsize,labelfont=sf,textfont=sf]{subfig}
\usepackage{booktabs}
\usepackage{textcomp}
\usepackage{multirow} 
\usepackage{array} 
\usepackage{wrapfig}
\usepackage{stfloats}
\usepackage{url}
\usepackage{verbatim}
\usepackage{graphicx}
\usepackage{cite}
 \usepackage{algorithm}
 \usepackage{algorithmicx}
 \usepackage{color} 

\usepackage{hyperref}
 \usepackage{algpseudocode}
\begin{document}

\title{ Hierarchical Error Assessment of CAD Models for Aircraft Manufacturing-and-Measurement}

\author{
 Jin Huang, Honghua Chen and Mingqiang Wei,~\IEEEmembership{Senior Member,~IEEE}
\thanks{Jin Huang and Mingqiang Wei are with School of Computer Science and Technology, Nanjing University of Aeronautics and Astronautics, Nanjing, China, and also with School of Artificial Intelligence, Taiyuan University of Technology, Taiyan, China (e-mail: jinhuang.nuaa@gmail.com, mingqiang.wei@gmail.com).} 
\thanks{Honghua Chen is with the College of Computing and Data Science,
Nanyang Technological University (NTU),
Singapore (e-mail: honghua.chen@ntu.edu.sg).}
}

\markboth{Journal of \LaTeX\ Class Files,~Vol.~14, No.~8, August~2021}%
{Shell \MakeLowercase{\textit{et al.}}: A Sample Article Using IEEEtran.cls for IEEE Journals}


\maketitle
\begin{abstract} 
The most essential feature of aviation equipment is high quality, including high performance, high stability and high reliability.
In this paper, we propose a novel hierarchical error assessment framework for aircraft CAD models within a manufacturing-and-measurement platform, termed HEA-MM. 
HEA-MM employs structured light scanners to obtain comprehensive 3D measurements of manufactured workpieces.
The measured point cloud is registered with the reference CAD model, followed by an error analysis conducted at three hierarchical levels: global, part, and feature. 
At the global level, the error analysis evaluates the overall deviation of the scanned point cloud from the reference CAD model. 
At the part level, error analysis is performed on these patches underlying the point clouds. 
We propose a novel optimization-based primitive refinement method to obtain a set of meaningful patches of point clouds. 
Two basic operations, splitting and merging, are introduced to refine the coarse primitives.
At the feature level, error analysis is performed on circular holes, which are commonly found in CAD models. 
To facilitate it, a two-stage algorithm is introduced for the detection of circular holes. First, edge points are identified using a tensor-voting algorithm. Then, multiple circles are fitted through a hypothesize-and-clusterize framework, ensuring accurate detection and analysis of the circular features.
Experimental results on various aircraft CAD models demonstrate the effectiveness of our proposed method.
\end{abstract}


\section{Introduction}
\label{introduction}
\IEEEPARstart{I}{n} the field of aerospace engineering, aircraft manufacturing necessitates a focus on safety, efficiency, and accuracy. 
Central to this process is computer-aided design (CAD) models, which constitute the foundational components of aircraft parts.
Any discrepancies or errors within the manufacturing process of CAD models may result in significant production delays and even compromise the safety of aircraft.
It is critical to ensure the manufacturing accuracy of these CAD models. 

\begin{figure*}[htbp]
	\centering
	\includegraphics[width=0.95\linewidth]{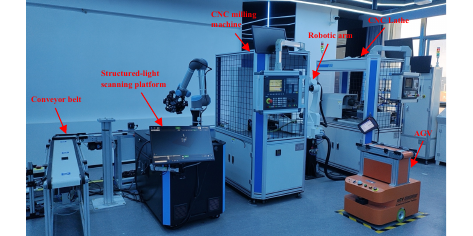}
	\caption{Our proposed manufacturing-and-measurement platform. 
    A robotic arm selects raw materials and feeds them into the manufacturing module, processing them into workpieces. 
    Then, the robotic arm moves the finished workpieces back to a conveyor belt, transferring them to the measurement module. 
    A light scanner is used to capture 3D point clouds of the objects. 
    Upon evaluating the manufacturing error, an AGV transports the workpieces to either a disposal area or a secure area, depending on the severity of the error. }
 \label{fig:platform}
\end{figure*}

Numerous techniques have been proposed to measure the manufacturing errors in aircraft mechanical parts. 
Traditional efforts using coordinate measuring machines (CMM), provide high-precision measurements of airplane workpieces. 
However, they often depend on manual inspection, making them labor-intensive and susceptible to human error~\cite{swornowski2013critical}, limiting their further practicality and scalability. 
Additionally, they might encounter difficulties handling the complex intricacies of aircraft designs. 
In recent years, the rapid advancement of sensing technology has significantly developed measurement methodologies, providing sophisticated tools to identify inaccuracies within CAD models. 
High-precision laser/optical scanners~\cite{decker2020efficiently,nguyen2018comparison} are utilized to collect 3D information of objects, assessing manufacturing errors with greater accuracy. 
For example,  Decker et al.~\cite{decker2020efficiently}  propose an efficient registration method for aligning scanned point clouds with CAD models, and then evaluate the global deviations in manufactured shapes.
However, there are still a few challenges to measure the manufacturing errors better:
\begin{enumerate}
\item How can manufacturing and measurement be integrated into a unified process to enhance the overall quality and efficiency?
\item How to effectively analyze manufacturing errors in an automated manner?
\end{enumerate}

To address these challenges, we introduce a novel framework for hierarchical error assessment within a manufacturing-and-measurement platform specifically tailored for aircraft CAD models.
As shown in Figure~\ref{fig:platform}, the complete platform comprises several integrated components designed to facilitate efficient production and error assessment processes. 
First, a robotic arm selects raw materials and feeds them into the manufacturing module, where they are processed into workpieces. 
Following the subtractive manufacturing (i.e., cutting and milling) of aircraft mechanical parts, the robotic arm moves the finished workpieces back to a conveyor belt, transferring them to the measurement module. 
Here, a light scanner captures measurement surfaces (3D point clouds) of the objects.

After obtaining the complete 3D measurement surface of airplane mechanical parts, our framework analyzes manufacturing errors by comparing the data with the reference digital CAD model. 
If the manufacturing error falls within acceptable limits compared to design specifications, an automated guided vehicle (AGV) transports the workpiece to the storage area. 
Conversely, if the error exceeds tolerable limits, the workpiece is deemed invalid and is moved to the disposal area.
\textit{We have recorded the whole system in the supplementary video}.

We develop an Aircraft Manufacturing-and-Measurement Platform, which has worked at several aircraft component manufacturers, such as COMAC, and Aero Engine Corporation of China, etc. It involves registering the scanned point cloud with the reference CAD model and systematically analyzing the point cloud at three top-to-down levels.
Unlike previous methods that primarily focus on either global or local deviations of point clouds in mechanical parts \cite{decker2020efficiently,nguyen2018comparison}, our framework enables accurate error assessment, thereby improving the efficiency of aircraft inspection. 
At the global level, we focus on evaluating the overall deviations of the point cloud.
At the part level, the error assessment is conducted on the underlying patches.
At the feature level, we shift our target to circular features commonly existing in mechanical parts.
Our contributions are three-fold:
\begin{enumerate}
    \item We propose a hierarchical error assessment framework for aircraft CAD models within a manufacturing-and-measurement platform. To the best of our knowledge, it is the first time.
    \item We introduce a primitive refinement method that leverages the merging and splitting operations to obtain satisfied geometric primitives.
    \item We present a method for detecting circular holes, utilizing a voting-based edge extraction strategy and a multi-circle fitting scheme.
\end{enumerate}

\section{Related work}
\label{related work}
Given the large number of methods related to manufacturing error assessment, we mainly focus on the fields of point cloud registration, primitive detection, and multi-structure fitting.

\subsection{Point cloud registration}
Chen et al.~\cite{chen2019plade} present a registration method for point clouds with low overlapping.
They design a novel plane/line-based descriptor dedicated to establishing high-level structural correspondences between point clouds.
Zhou et al.~\cite{zhou2016fast} propose a fast algorithm for the global registration of partially overlapping point clouds.
Li et al.~\cite{li2024hybrid} present a coarse registration method for point clouds obtained from laser scanning and photogrammetric reconstruction.

Some learning-based point cloud registration methods~\cite{lu2019deepvcp,sarode2019pcrnet,yin2022deep, chen2022imlovenet} are also proposed.
Lu et al.~\cite{lu2019deepvcp} introduce DeepVCP, a novel end-to-end point cloud registration network.
Rather than searching the corresponding points among existing points, they generate them based on learned matching probabilities among a group of candidates.
Sarode et al.~\cite{sarode2019pcrnet} present a PCRNet, comparing PointNet features of the source and target point clouds to calculate the transformation matrix that best aligns them. 
Iterated Closest Point (ICP) is the most commonly used fine registration method of point clouds. 
In this paper, we adopt a coarse-to-fine strategy to register the scanned point cloud and the reference CAD model.

\subsection{Primitive detection}
The most commonly used primitive detection methods are random sample consensus (RANSAC)~\cite{fischler1981random} and its variants~\cite{schnabel2007efficient}, which are robust against noise and outliers.
The other popular type is the region growing-based methods~\cite{rabbani2006segmentation}, which iteratively propagate planar regions by boundary advancement. 
The main differences between existing region-growing methods arise from variations in their seed point initialization and region expansion criteria.
Yu et al.~\cite{Yu_2022_CVPR} propose a refinement approach incorporating five types of operations: merging, splitting, transfer, insertion, and exclusion, enhancing the planar segmentation results.

As deep learning techniques continue growing, various learning-based primitive detection methods~\cite{qi2017pointnet++,li2023surface,wang2020unequal} have been proposed.
PointNet++~\cite{qi2017pointnet++} extracts point-wise features for individual points, enabling the estimation of the planar primitive types and parameters.
Li et al.~\cite{li2023surface} present a novel surface and edge detection network (SED-Net) for accurate geometric primitive fitting of point clouds.
The key idea is to learn parametric surfaces and edges simultaneously that can be assembled into a regularized and seamless Computer-Aided Design (CAD) model in a single framework.
Overall, learning-based methods have shown limitations in handling real-world data, mainly because the training datasets differ from real scenarios as they lack defects and often consist of synthetic CAD models. 
In this paper, we introduce an optimization-based primitive detection method without relying on any training process.

\subsection{Multi-structure fitting}
The task of identifying underlying structures within given point clouds is inherently challenging due to unavoidable noises and outliers. 
Recent advancements have led to the development of several multi-structure fitting algorithms designed to seek multiple models that best represent the data~\cite{magri2016multiple,wang2018searching}.
Magri et al.~\cite{magri2016multiple} extend the RANSAC algorithm to fit multiple structures by framing the detection procedures as a set coverage problem.
Wang et al.~\cite{wang2018searching} propose a multi-structure fitting method that can effectively and efficiently estimate the number and the parameters of model instances in the data.
However, these methods are often unsatisfactory when applied to real scanned data, particularly in more complex scenarios.

Abdual et al.~\cite{nurunnabi2018robust} present an efficient algebraic circle fitting method based on Robust Principal Component Analysis (RPCA).
Their method demonstrates high accuracy in estimating circle parameters, even in the presence of missing data and severe outliers.
Wang et al.~\cite{wang2016shape} introduce a subspace-clustering-based multi-structure detection method, capable of identifying multiple structures within LiDAR point clouds despite significant outliers. 
However, their method is primarily designed for plane detection.
In this paper, we propose a novel method for detecting multiple underlying circles in real-scanned point clouds to analyze manufacturing errors.

\section{Methodology}
\label{methodology}

\begin{figure*}[htbp]
	\centering
	\includegraphics[width=0.92\linewidth]{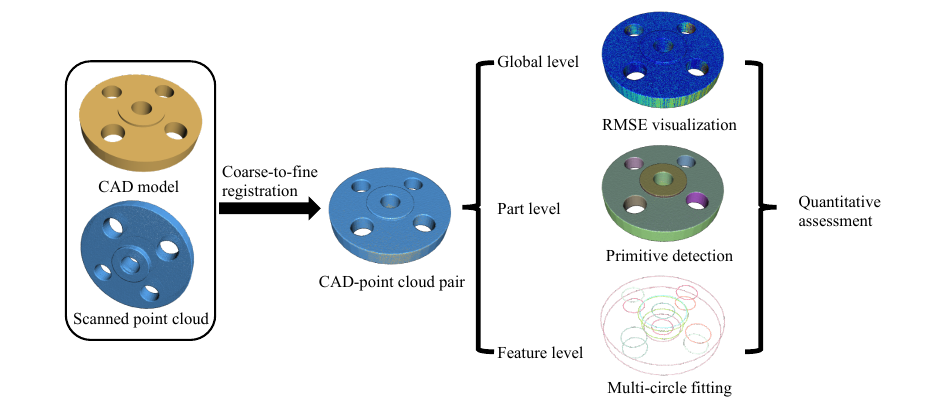}
	\caption{The pipeline of our method. Taking a scanned point cloud and the reference CAD mesh as input, we first register them together. Then three top-to-down level error analyses are conducted on the point cloud.  }
 \label{fig:pipeline}
\end{figure*}

\subsection{Overview}
Our framework takes a scan of an airplane part and the reference CAD model as input.
The goal is to produce an analysis report on manufacturing errors. The overall pipeline is depicted in Figure~\ref{fig:pipeline}.
Specifically, the scanned point cloud is registered with the reference CAD model. Then, an error assessment is conducted at three top-to-down levels on the registered point cloud as follows:

\begin{itemize}
  \item \textbf{Global level:} The assessment evaluates the overall deviation of the scanned point cloud from the reference model. 
  \item \textbf{Part level:} We introduce a novel primitive refinement method that divides the input point cloud into a set of meaningful patches used for part-level error assessment. 
  \item \textbf{Feature level:} Circular holes are fitted from the point cloud using a new multi-circle fitting method, which is used for feature-level error analysis.
\end{itemize}
This multi-level assessment framework allows for a comprehensive comparison with the design specifications of aircraft CAD models.

\subsection{CAD-point cloud registration}
Given a scanned point cloud and the reference CAD model, the goal is to align them accurately.
Since the inherent imperfections commonly found in raw point clouds, pre-processing~\cite{chen2022repcd} becomes imperative to improve the data quality.
We employ the method proposed by Chen et al.~\cite{chen2019multi} to remove noises and outliers.
After filtering out imperfect points, the remaining data forms the source point cloud. Subsequently, a point set is resampled from the CAD model to serve as the target point cloud. This process results in a pair of point clouds, which is utilized for subsequent point cloud registration.

We adopt a coarse-to-fine strategy to register this point cloud pair.
These two point clouds are coarsely registered by using the Fast Global Registration (FGR)~\cite{zhou2016fast} method.
The initial alignment is refined through a fine registration method, specifically the Iterative Closest Point (ICP) algorithm~\cite{besl1992method}.
After registering the scanned point cloud with the reference CAD mode, we analyze manufacturing errors using a hierarchical framework.
It comprises three levels of error assessment:  global level, part level, and feature level.

\subsection{Global level}
For the global level error, we utilize the root mean square error (RMSE) to assess the overall deviation between a point cloud and the reference CAD model $M$ as
\begin{equation}
E_{global} = \sqrt{\frac{1}{n}\sum_{i=1}^{n} dist^2(v_i,M)},
\label{equ:global_metric}
\end{equation}
where  $dist(v_i,M)$ denotes the distance from point $v_i$ to the reference CAD model $M$.

\subsection{Part level}
For the part level error, we compute the distance of each patch to the reference CAD model.
A patch is defined as a group of points sharing similar characteristics. 
Given that CAD models typically comprise basic primitives such as planes, spheres, cylinders, and cones, we divide the registered point cloud into several homologous patches. 
Inspired by the approach in~\cite{Yu_2022_CVPR}, which was specifically designed for planar surfaces, we extend it to refine non-planar primitives.

Our algorithm takes a point cloud as input and outputs a set of clusters of inlier points. 
First, a coarse primitive detection result is obtained using a region-growing-based method, which employs curvature as the metric to measure point deviation. 
Once the label of each point is assigned, the optimal primitive type and corresponding parameters
\textit{s} are determined. 
Here, we utilize four basic primitive types: plane, sphere, cone, and cylinder.
Besides, to refine the extracted primitives, we introduce two fundamental operations: splitting and merging, minimizing an energy function.

\textbf{Energy function.} The quality of a primitive configuration $x$ is measured by an energy $E$, formulated as
\begin{equation}
E(x)=\omega_f E_f(x) + \omega_s E_s(x)+ \omega_c E_c(x),
\label{equ:energy_function}
\end{equation}
 where $E_f(x)$, $E_s(x)$, and $E_c(x)$ denotes the fidelity term, simplicity term, and completeness term, respectively.
 $w_f$, $w_s$, and $w_c$ are the corresponding weights to balance these terms.

\textbf{Fidelity term} measures how well the primitives fit an input point cloud,  defined as  the average distance between inlier points and the corresponding primitive, formulated as
 \begin{equation}
 E_f(x) = \frac{1}{n_x}\sum_{p\in P}\sum_{v_i\in p} d_\epsilon (v_i,p),
\label{equ:fidelity}
\end{equation}
where $n_x$ is the total number of inlier points in the primitive configuration $x$, and $d_\epsilon (v_i,p)$ is the distance from point $v_i$ to the primitive $p$.

\textbf{Simplicity term} favors the final result with a low number of primitives, written as
\begin{equation}
 E_s(x) = \frac{\left| P \right|}{n_\sigma},
\label{equ:simplicity}
\end{equation}
where $\left| P \right|$ is the number of primitives, and $n_\sigma$ denotes the maximal number of detected primitives, simply computed as the ratio of the number of input points $n$ to the minimal number of points $\sigma$ per primitive.

\textbf{Completeness term} encourages the refined primitives with a high ratio of inlier points, defined as
\begin{equation}
 E_c(x) = 1-\frac{n_x}{n},
\label{equ:completeness}
\end{equation}

\begin{wrapfigure}{r}{0.23\textwidth}
  \centering
  \includegraphics[width=0.23\textwidth]{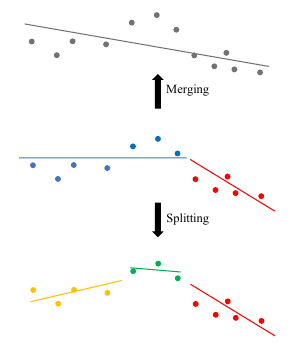}
\end{wrapfigure}
\textbf{Basic operations.}
We employ merging and splitting operations to refine the set of coarse primitives.
The splitting operation, as depicted in the bottom row of the right figure, divides a single primitive into two distinct primitives.
Specifically, the inlier points are divided into two groups.
First, the two farthest points on each side of a primitive are identified.
Next, the other inlier points are associated with these two points depending on their geometry proximity. 
Lastly, two new primitives are computed for each group of points.
Conversely, the merging operation combines two adjacent primitives into a single new primitive, as shown in the bottom row of the right figure.
To simplify the overall process, we assume that the type of a primitive remains unchanged before and after the operation.

\textbf{Exploring mechanism.} Staring from a good configuration $x$, the main idea is to compute the energy decreased by the splitting or merging operations that involve re-assigning inlier and outlier points for the primitives.
We sort the energy reduction of all possible operations in a priority queue, ranking the operations that help improve primitive refinement.
The whole exploration stops when no more operations help decrease the energy. 
This stop condition guarantees the exploration mechanism to converge quickly. 
The pseudo-code of our algorithm is summarized in Algorithm~\ref{alg:primitive_refinement}.

After minimizing the overall energy $E$ in Equation~\ref{equ:energy_function}, we can obtain a set of refined primitives underlying the point cloud. 
The part level error is defined as the average/maximum distance of the segmented patches to the reference CAD model, computed as
\begin{equation}
    \begin{split}
        E_{\text{part\_avg}} &= \frac{\sum_{i=1}^{n} \text{dist}(p_i, M)}{\left| P \right|}, \\
        E_{\text{part\_max}} &= \max \{\text{dist}(p_i, M)\},
    \end{split}
    \label{equ:part_metric}
\end{equation}
where dist$(p_i,M)$ denotes the distance from primitive $p_i$ to the reference CAD model $M$.

\begin{algorithm}
\caption{   Exploration mechanism of primitive refinement}
\begin{algorithmic}[1]
    \State \textbf{Input:} $\text{A set of coarse primitives}$  
     \State \textbf{Output:} $\text{A set of  refined primitives}$ 
    \Repeat
    \While{operation i  decreases energy E} 
        \State $\text{Update configuration x by operation i  }$ 
        \State $\text{Update Q}$ 
    \EndWhile
    \Until{no update decreases {E} any more}

\end{algorithmic}
\label{alg:primitive_refinement}
\end{algorithm}

\subsection{Feature level}
At the feature level, we focus on assessing the manufacturing accuracy of circular holes, commonly existing in aircraft CAD models.
It is defined as the average/maximum deviations of centroid positions and the circle radii.
To accurately identify the circular holes underlying the point cloud, we utilize a tensor voting based method to extract the edge points, followed by a novel multi-circle fitting scheme (MCFS).

\subsubsection{Edge point detection} 
The original tensor voting~\cite{park2012multi} algorithm is proposed for feature detection of unstructured point clouds through transferring tensors between neighboring points.
It mainly consists of three steps, tensor coding, tensor voting, and tensor decomposition.
In our paper, we use this algorithm to detect edge points underlying an unstructured point cloud.
Based on the observation that the points belonging to the same primitive often share similar curvature, we introduce a new weight to refine the voting scheme.
The tensor $T_i$ of point $v_i$ is defined as
\begin{equation}
    T_i=\sum_{j\in N(i)} \mu_j \Delta \kappa_j(I_3- \frac{v_j v_j^T}{\| v_j v_j^T\|})
    \label{equ:refined_tensor}
\end{equation}
where $\Delta \kappa_j$ is defined as $\|\kappa_i-\kappa_j\|$, decipting the deviation of point curvature. 
$N(i)$ is the set of the points around $v_i$ in $\epsilon$-radius sphere.
The vector $q_j=\|v_i-v_j\|$, and $I_3$ is the identity matrix, $\mu_j =exp({-\frac{\|q_j\|}{\sigma^2}})$ is the gaussian function, $\sigma$ is fixed to be 0.5. 
$T_{i}$ is a semi-define matrix and can be written as
\begin{equation}
    T_i=\lambda_1 e_1 e_1^T +\lambda_2 e_2 e_2^T +\lambda_3 e_3 e_3^T 
    \label{equ:tensor_decom}
\end{equation}
where $e_1,e_2,e_3$ are the eigenvectors, and $\lambda_1\geq \lambda_2 \geq\lambda_3 \geq 0$ are its corresponding eigenvalues.
A pointwise feature $W_i$ is defined by using these eigenvalues as 
\begin{equation}
    W_i=\frac{\lambda_2+\lambda_3}{\lambda_1}
    \label{equ:weight}
\end{equation}
In our experiments, we empirically set $W_i \geq$ 1.05, which indicates that the point belongs to the group of edge points.

\subsubsection{Multi-circle fitting} After getting the edge points, we can detect multiple circles and estimate their parameters (e.g., radius and centroid).
We formulate it as a multi-label classification problem, solved by a novel multi-circle fitting scheme (MCFS)  based on Multi-class Cascaded T-linkage (MCT)~\cite{magri2019fitting}.
The method follows a \textit{hypothesize-and-clusterize} framework: the input data is firstly represented by a set of model hypotheses, and then a hierarchical clustering is performed to generate a meaningful partition of data.
The pseudo-code of our algorithm is summarized in Algorithm~\ref{alg:mcf}.

\textbf{Hypothesis generation.} This method starts with generating a set of hypotheses $H = \{ h_1,h_2,\dots,h_m\}$ from a minimum number of points, in our circle-fitting case, it refers to 3.
Since the number of hypotheses is often much more than the number of circles, we use the mean shift algorithm~\cite{comaniciu2002mean} to reduce the number of hypotheses. 

\textbf{Matrix calculation.} 
The residual matrix $R$ represents the distances from the data points to the model hypotheses, evaluating their correlation.
The $n \times m $ matrix is calculated as:
$ R=\{ d_{i,j}| i=1,2,\dots,n; j=1,2,\dots,m \}$, where $d_{i,j}$ is the distance from point $v_i $ to the model hypothesis $h_j$.
$n$ and $m$ are the number of points and model hypotheses, respectively.

\textbf{Labeling.} 
Using the residual matrix $R$, we can classify the points through hierarchical clustering. 
Points that belong to the same circle tend to exhibit similar residual values. 
The smaller the point-circle distance, the
greater the likelihood of the belonging model.
Thus, the circle instances are separated through hierarchical clustering from the residual matrix $R$.

\begin{figure}[ht]
	\centering
	\includegraphics[width=0.95\linewidth]{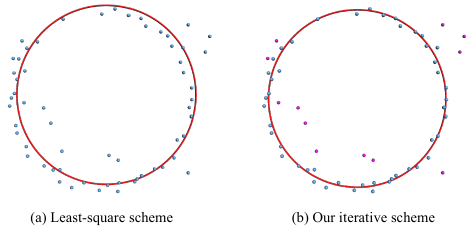}
	\caption{Comparison of circle fitting results.  (a) Least-square scheme. (b) Our iterative scheme. The purple points in (b) denote the outlier points. }
 \label{fig:iterative_fitting}
\end{figure}

\textbf{Outlier rejection.}
After getting a set of clustered points, we can estimate the parameters of a circle from the data.
Instead of simply using the least-square algorithm to fit a circle, we utilize an iterative scheme to mitigate the effect of outliers and noises.
We start from a circle fit from all input points and compute the radius and centroid.
For the outlier points whose distance to the circle is larger than a threshold $\epsilon$, we discard these points and update the circle parameters.
The iteration is stopped if the variation in parameters is smaller than a threshold $\eta$ or the iteration number reaches a maximum value $I_t$. 
The fitted circle in the last iteration is used as the final result.
As shown in Figure~\ref{fig:iterative_fitting}, our iterative scheme generates better fitting results, thanks to the outlier rejection mechanism.

\textbf{Refinement.} A final refinement step is used to update the circle parameters.
Given that some hypothesis circles are close to each other, a circle model likely corresponds to more than one hypothesis model.
To address this issue, we solve a set coverage problem~\cite{magri2016multiple} using a greedy algorithm, selecting the most suitable models in a decreasing order.

\begin{algorithm}
\caption{Multi-circle fitting scheme} 
\begin{algorithmic}[1]
    \State \textbf{Input:} $\text{Edge points }  T $  
    \State \textbf{Output:} $\text{Fitted circles }  C, \text{Labeling } L$ 
    \State \textbf{Initialization:} $T' =T$
    \While{obtain all circles} 
        \State $H_0 \gets \text{HypothesisGeneration}(T');$ 
        \State $\mathcal{R}\gets \text{MatrixCalculation}(T',H_0);$ 
        \State $(L_i,C_i) \gets \text{Labeling}(T', \mathcal{R});$ 
        \State $(T',L_i,C_i)\gets \text{OutliersRejection}(T',L_i);$
        \State $(L,C) \gets \text{CircleInstanceUpdate}(L_i,C_i);$
    \EndWhile

\end{algorithmic}
\label{alg:mcf}
\end{algorithm}
After obtaining the fitted circles and  determining their parameters, we compute feature level error,  defined as the average/maximum deviation of centroid positions and  circle radii with ground truth from the reference CAD model as
\begin{equation}
    \begin{aligned}
        E_{\text{comp\_radius\_max}} &= \max \{|r_i - r_i^*| \} \\
        E_{\text{comp\_radius\_avg}} &= \frac{\sum_{i=1}^{n} |r_i - r_i^*|}{N} , \\
        E_{\text{comp\_centroid\_max}} &= \max \{|o_i - o_i^*| \}\\
        E_{\text{comp\_radius\_avg}} &= \frac{\sum_{i=1}^{n} |o_i - o_i^*|}{N} , \\
    \end{aligned}
    \label{equ:com_metric}
\end{equation}
where $r_{i}$ and $r_{i}^*$ are the estimated radius and the corresponding ground truth radius of the circle $C_i$,  $o_{i}$ and $o_{i}^*$ are the estimated centroid and the corresponding ground truth centroid of the circle $C_i$. $N$ is the number of circle instances.

\begin{figure*}[ht]
	\centering
	\includegraphics[width=1\linewidth]{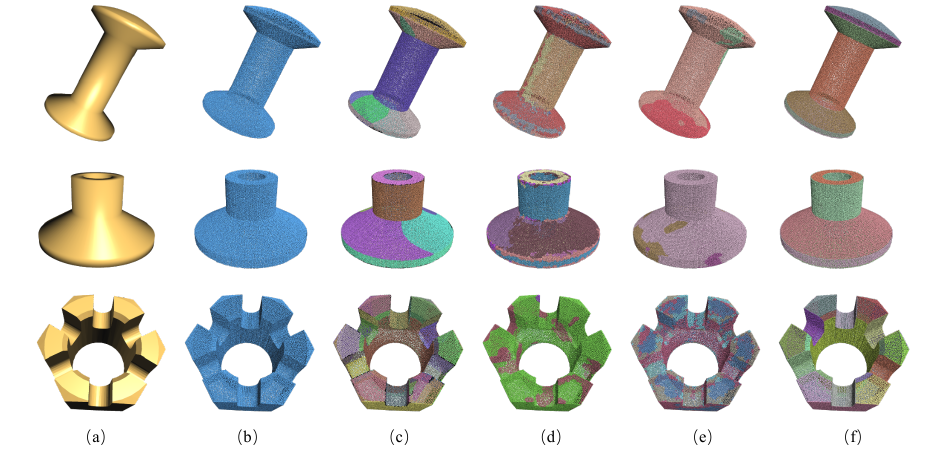}
	\caption{The comparison of our method with other state-of-the-art methods.
 (a) Reference CAD mesh.
 (b) Scanned point cloud.
 (c) Ransac~\cite{schnabel2007efficient}.
 (d) Pointnet++~\cite{qi2017pointnet++}.  (e) HPNet~\cite{yan2021hpnet}.
 (f) Our result.}
 \label{fig:comparison_primitive_detection}
\end{figure*}

\begin{figure*}[ht]
	\centering
	\includegraphics[width=0.90\linewidth]{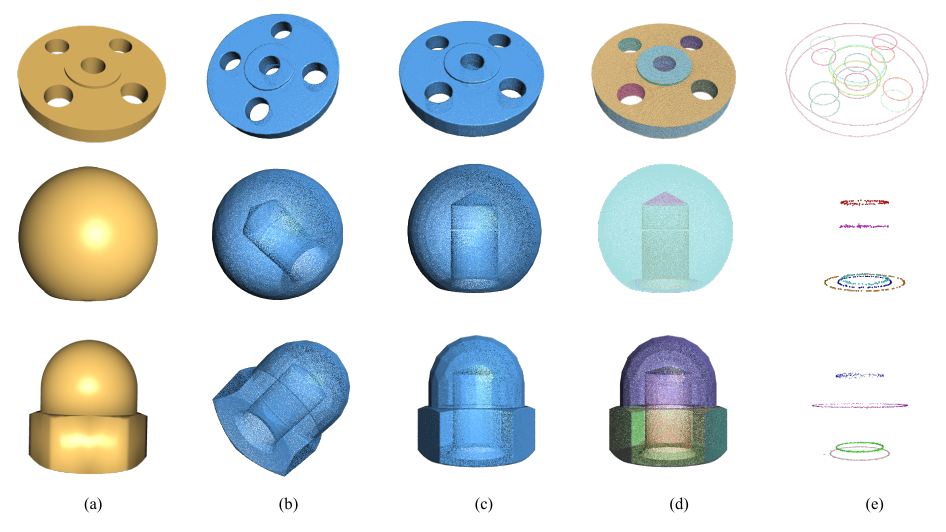}
	\caption{The error analysis results of our method.
  (a) Reference CAD model. (b) Scanned point cloud. (c) Registered point cloud. (d) Primitive detection result. (e) Multi-circle fitting result.}
 \label{fig:results}
\end{figure*}

\section{Results and discussion}
\label{experiment}
We have tested our method on a variety of airplane CAD models. The point clouds are collected by using CREAFORM MetraScan scanner.
Our algorithms are implemented in C++ and Matlab.
All experiments are conducted on a desktop PC with an AMD Ryzen5 5600G CPU and 16GB memory.
\textit{More results are provided in the supplementary material.}
\begin{table*}[ht]
\centering
\caption{Manufacturing error (in millimeters) of aircraft CAD models.}
\label{table:statistical_analysis}
\begin{tabular}{|c|c|c|c|c|c|c|c|c|}
  \hline
  & \multicolumn{1}{c|}{Registration error} & \multicolumn{1}{c|}{Global level} & \multicolumn{2}{c|}{Part level} & \multicolumn{4}{c|}{Feature level} \\ \cline{2-9}
  & $E_{\text{reg}}$ & $E_{\text{global}}$ & $E_{\text{part\_avg}}$ & $E_{\text{part\_max}}$ & $E_{\text{comp\_radius\_avg}}$ & $E_{\text{comp\_radius\_max}}$ & $E_{\text{comp\_centroid\_avg}}$ & $E_{\text{comp\_centroid\_max}}$ \\ \hline
  (1) & 0.0076 & 0.0902 & 0.0651 & 0.0773 & 0.0625 & 0.0941 & 0.0796 & 0.0824 \\ \hline
  (2) & 0.0064 & 0.0805 & 0.0732 & 0.0761 & 0.0814 & 0.0837 & 0.0703 & 0.0729 \\ \hline
  (3) & 0.0049 & 0.0751 & 0.0593 & 0.0628 & 0.0699 & 0.0715 & 0.0548 & 0.0573 \\ \hline
\end{tabular}
\end{table*}
\subsection{Comparison}
\subsubsection{Primitive detection}
We compare our primitive detection method with state-of-the-art methods, including the traditional RANSAC~\cite{schnabel2007efficient}, two learning-based methods, Pointnet++~\cite{qi2017pointnet++} and HPNet~\cite{yan2021hpnet}.

For a simple CAD model like a rivet in Figure~\ref{fig:comparison_primitive_detection}, all four methods can output a meaningful decomposition of the point cloud. 
However, Ransac misclassifies some primitives and learning-based methods run into problems with the edge regions.
The main reason is that the normals of these points are hard to be accurately computed.
Thanks to the exploration mechanism of primitive refinement,  our method can accurately obtain a set of primitives from the point cloud without any training process.
The learning-based framework is hard to obtain satisfied primitives from real scanned point clouds. 
This type of method is often trained on the ABC dataset~\cite{koch2019abc}, which does not take into account the noises or outliers in real scans.
Since there is a lack of benchmark of real scanned CAD models, new training is not applicable. 
\begin{figure}[htbp]
	\centering
	\includegraphics[width=1.0\linewidth]{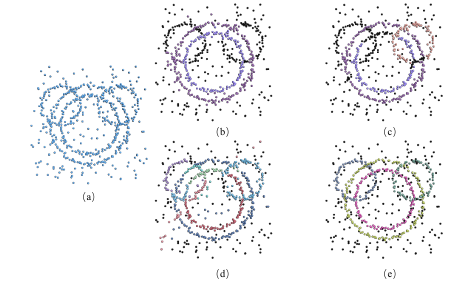}
	\caption{  Comparisons of multi-circle fitting results on a 2D synthetic example.  (a) Input points.   (b) KF~\cite{chin2009robust}. (c) T-linkage~\cite{magri2019fitting}. (d) MSH~\cite{wang2018searching}. (e) Our result. The points with different colors belong to different clusters, and black indicates outliers. }
 \label{fig:circle_fitting1}
\end{figure}

\begin{figure}[htbp]
	\centering
	\includegraphics[width=0.95\linewidth]{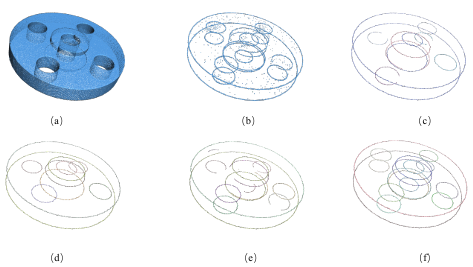}
	\caption{Comparisons of multi-circle fitting results on a scanned point cloud.  (a) Scanned point cloud. (b) Edge points. (c) KF~\cite{chin2009robust}. (d) T-linkage~\cite{magri2019fitting}. (e) MSH~\cite{wang2018searching}.
      (f) Our result. }
 \label{fig:circle_fitting2}
\end{figure}

\subsubsection{Multi-circle fitting}
We compare our multi-circle fitting method with other state-of-the-art methods, including KF~\cite{chin2009robust}, T-linkage~\cite{magri2019fitting}, and  MSH~\cite{wang2018searching}.

We test a 2D synthetic example consisting of 4 circles corrupted with 10\% noises and 50\% outliers.
As shown in Figure~\ref{fig:circle_fitting1}, our method accurately detects all 4 circles from the points.
KF~\cite{chin2009robust} and T-linkage~\cite{magri2019fitting} only detect 2 or 3 circles. 
The fitting results of MSH~\cite{wang2018searching} are over-classified, and some edge points are misjudged.
As shown in Figure~\ref{fig:circle_fitting2}, we test on a real scanned point cloud.
Our method is the only one that can detect all 16 circular holes. The other methods fail to detect small circular holes, or some edge points are misjudged.

\subsection{Discussion}
The point cloud registration error is defined as the average distance between the registered scanned point cloud and the sampled point cloud from the CAD mesh model.
As shown in Table~\ref{table:statistical_analysis}, the registration error is often one magnitude smaller than the manufacturing errors (namely three-level errors).

After the registration, we conduct three levels of error analysis on the CAD models, the visual results are shown in Figure~\ref{fig:results}.
First, RMSE is employed to compute the global level error, quantifying the deviation of the manufactured workpiece from the CAD models.
The smaller the value is, the more accurately the workpiece is manufactured.
Then, we compute the part-level error.
At this level, the point cloud is decomposed into smaller patches, in this way we can analyze the manufacturing error in a smaller region.
Lastly, we compute the feature level error. 
Our goal is to evaluate the manufacturing error of circular holes, commonly existing in aircraft CAD models.
The maximum/average deviations of radii and centroids of fitted circle instances are calculated.

The overall running times of the steps are recorded in Table~\ref{table:running_time}.
Note that the complete platform involves the process of manufacturing and measuring, but we only record the running time of the proposed algorithms apart from the scanning process, often within a few minutes. 

\begin{table}[ht]
\centering
\caption{The running time (in seconds) of main steps for aircraft CAD models in Figure~\ref{fig:results}.}
\label{table:running_time}
\begin{tabular}{|c|>{\centering\arraybackslash}m{1.4cm}|c|c|c|c|c|c|}
  \hline
  & \begin{tabular}[c]{@{}c@{}}Registration\end{tabular} & \shortstack{Global-\\level} & \shortstack{Part-\\level} & \shortstack{Feature-\\level} & \shortstack{Total \\ time} \\ \hline
  (1) & 2.3 & 1.2 & 23.4 & 26.2  & 53.1\\ \hline
  (2) & 1.6 & 0.9& 17.8 & 12.9  &33.2
  \\ \hline
  (3) & 3.5 & 1.4 & 28.1 & 10.9 &43.9
  \\ \hline
\end{tabular}
\end{table}
\subsection{Parameters}
Our method has several parameters that are empirically fixed for the experiments.
The weights in Equation~\ref{equ:energy_function} are set to $\omega_f=0.6$, $\omega_s=0.25$, and $\omega_c=0.15$, respectively.
The inline distance $\epsilon$ is set to 0.1.
In the multi-circle fitting step, the threshold $\eta$ for parameter variations is set to 0.03, and the maximum iteration number $I_t$ is set to 150.
\subsection{Limitations and future work}
At the part level, to obtain a set of meaningful patches, we assume that the aircraft CAD models consist of four types of primitives, namely planes, spheres, cylinders, and cones. 
It might fail to get satisfied segmentation results due to the B-spline/NURBS surfaces, or other more complex shapes.
At the feature level, we merely analyze circular holes, and a promising direction is to extend our method to include more basic features, such as square holes.

\section{Conclusion}
\label{conclusion}
We have introduced a novel error assessment framework of airplane CAD models with a manufacturing-and-measurement platform.
After registering the scanned point cloud with a reference CAD model,  we conduct three top-to-down levels of analysis of manufacturing error, global, part, and feature.
An optimization-based primitive refinement scheme is proposed to better obtain the patches underlying the point clouds.
A new multi-circle fitting method is presented to detect the circle structures in the point clouds.
Extensive experiments on aircraft CAD models show the effectiveness of our method.

\bibliographystyle{IEEEtran}
\bibliography{sources/references.bib}

\newpage
\newpage
This supplementary material presents additional experiments with (i) results of primitive detection from various point clouds, (ii)  examples of multi-circle fitting, and (iii) error analysis of CAD models.
\begin{figure*}[ht]
	\centering
	\includegraphics[width=1\linewidth]{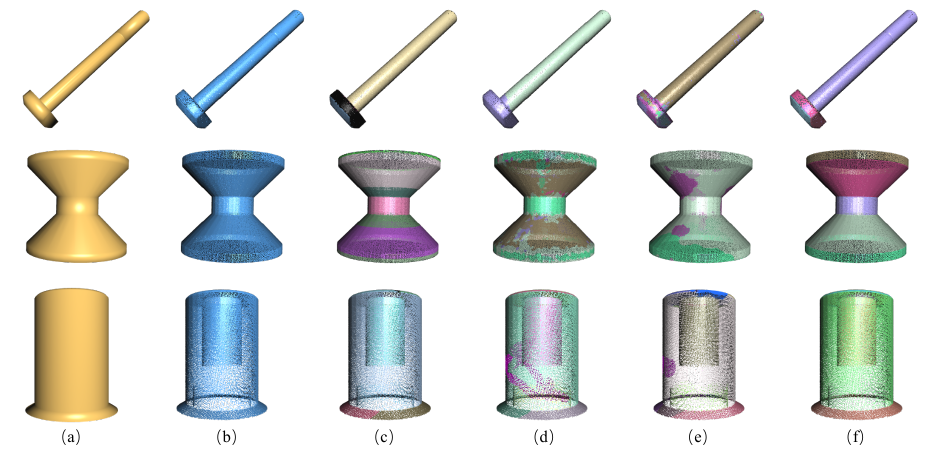}
	\caption{The comparison of our method with other state-of-the-art methods.
 (a) Reference CAD mesh.
 (b) Scanned point cloud.
 (c) Ransac~\cite{schnabel2007efficient}.
 (d) Pointnet++~\cite{qi2017pointnet++}.  (e) HPNet~\cite{yan2021hpnet}.
 (f) Our result.}
 \label{fig:comparison_primitive_results_sup}
\end{figure*}
\begin{figure*}[ht]
	\centering
	\includegraphics[width=0.95\linewidth]{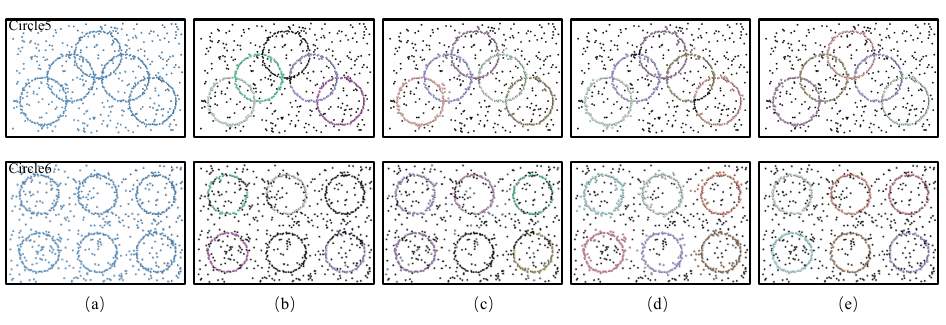}
	\caption{  Comparisons of multi-circle fitting results on 2D synthetic examples.  (a) Input points.   (b) KF~\cite{chin2009robust}. (c) T-linkage~\cite{magri2019fitting}. (d) MSH~\cite{wang2018searching}. (e) Our result. The points with different colors belong to different clusters, and black indicates outliers. }
 \label{fig:comparison_circle_results_sup}
\end{figure*}
\section{Primitive detection} 
This paper mainly focuses on primitive detection on real scanned point clouds, and there is no ground truth of the detection results. 
Therefore, we show the visual comparisons with state-of-the-art methods, including the traditional RANSAC~\cite{schnabel2007efficient}, two learning-based methods, Pointnet++~\cite{qi2017pointnet++} and HPNet~\cite{yan2021hpnet}.

For the mechanical CAD models depicted in Figure~\ref{fig:comparison_primitive_results_sup}, both our method and the RANSAC algorithm produce meaningful decompositions of the point cloud. 
However, the RANSAC algorithm frequently results in over-segmented or under-segmented primitives. 
In contrast, our method, utilizing merge and split operations, achieves more accurate primitive segmentation. 
Learning-based methods encounter difficulties with edge points, primarily due to the challenges in accurately computing the normals at these points.

\begin{figure*}[ht]
	\centering
	\includegraphics[width=0.95\linewidth]{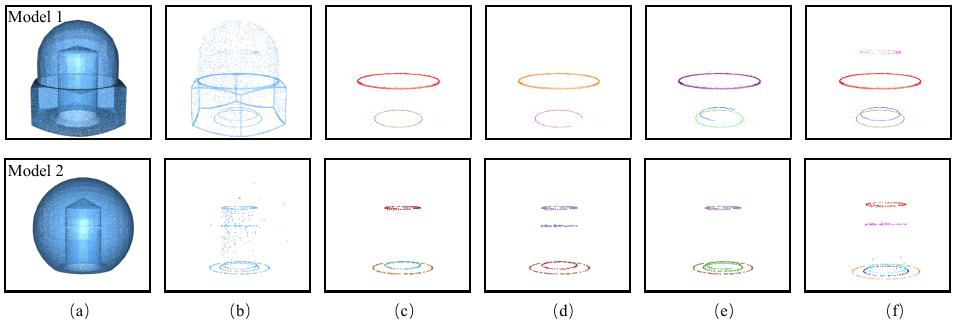}
	\caption{Comparisons of multi-circle fitting results on a scanned point cloud.  (a) Scanned point cloud. (b) Edge points. (c) KF~\cite{chin2009robust}. (d) T-linkage~\cite{magri2019fitting}. (e) MSH~\cite{wang2018searching}.
      (f) Our result. }
 \label{fig:circle_fitting2}
\end{figure*}

\section{Multi-circle fitting}
We compare our multi-circle fitting method with other state-of-the-art methods, including KF~\cite{chin2009robust}, T-linkage~\cite{magri2019fitting}, and  MSH~\cite{wang2018searching}.

We test the multi-circle fitting methods on 2D synthetic examples, as shown in Figure~\ref{fig:comparison_circle_results_sup}.
KF~\cite{chin2009robust} and  T-linkage~\cite{magri2019fitting} fail to detect all circles, due to the existing noises and outliers.
Meanwhile, both MSH~\cite{wang2018searching} and our method detect all circles.
To quantitatively evaluate multi-circle fitting results, Misclassification Error (ME) is used to measure the percentage of misclassified points with respect to the ground-truth models, defined as
\begin{equation}
    ME=\frac{\text{\# mislabeled data points}}{ \text{\# data point} } \times 100\%
\end{equation}
A lower value indicates a better result.
As shown in Table~\ref{tab:comparison_me}, our method detects the most accurate circles with the lowest percentage of mislabeled points.
\begin{table}[ht]
\centering
\begin{tabular}{|l|l|c|c|c|c|}
\hline
{Data} & {ME} & {KF~\cite{chin2009robust}} & { T-linkage~\cite{magri2019fitting}} & {MSH~\cite{wang2018searching}} & {Ours} \\
\hline
\multirow{2}{*}{circle5} & mean   & 30.4     &  11.2    &   15.6   & \textbf{7.8 }    \\
                        & median &   30.8   & 11.6     &   15.9   &   \textbf{8.1 }  \\
\hline
\multirow{2}{*}{circle6}  & mean   &  41.7    &  31.1    & 12.6     &  \textbf{9.2}   \\
                        & median &    42.1  &   31.5   &   12.9   &   \textbf{9.4}   \\
\hline
\end{tabular}
\caption{Misclassification Error (ME\%) for synthetic data on multi-circle fitting task. The mean and the median results are the average and the middle value of five runs. }
\label{tab:comparison_me}
\end{table}

We also test on real scanned point clouds, as shown in Figure~\ref{fig:circle_fitting2}.
The number of false negatives and false negative cases~\cite{dai2021aircraft} is used as the evaluation metric to 
 estimate the results, denoted as $f_n$ and $f_p$, as reported in Table~\ref{tab:comparison_fp_fn}.
Our method demonstrates superior accuracy in multi-circle detection, characterized by the lowest false negative ($f_n$) and false positive ($f_p$). 
In contrast, the other three methods either misclassify points or fail to detect all circles.

\begin{table}[ht]
\centering
\begin{tabular}{|c|c|c|c|c|c|c|c|c|c|}
\hline
\multirow{2}{*}{Data} & \multicolumn{2}{c|}{KF~\cite{chin2009robust}} & \multicolumn{2}{c|}{T-linkage~\cite{magri2019fitting}} & \multicolumn{2}{c|}{MSH~\cite{wang2018searching}} & \multicolumn{2}{c|}{Ours} \\ \cline{2-9}
 & $f_n$ & $f_p$ & $f_n$ & $f_p$ & $f_n$ & $f_p$ & $f_n$ & $f_p$ \\ \hline
Model1 & 0 & 2 & 0 & 2 & 0 & 1 & 0 & 0 \\ \hline
Model2 & 0 & 2 & 0 & 1 & 1 & 2 & 0 & 0 \\ \hline
\end{tabular}
\caption{Comparison of false negatives ($f_n$) and false positives ($f_p$) across different methods.
The $f_n$ and $f_p$ denote the number of errors and misses.}
\label{tab:comparison_fp_fn}
\end{table}

\section{Error analysis}
\begin{figure*}[t]
	\centering
	\includegraphics[width=0.95\linewidth]{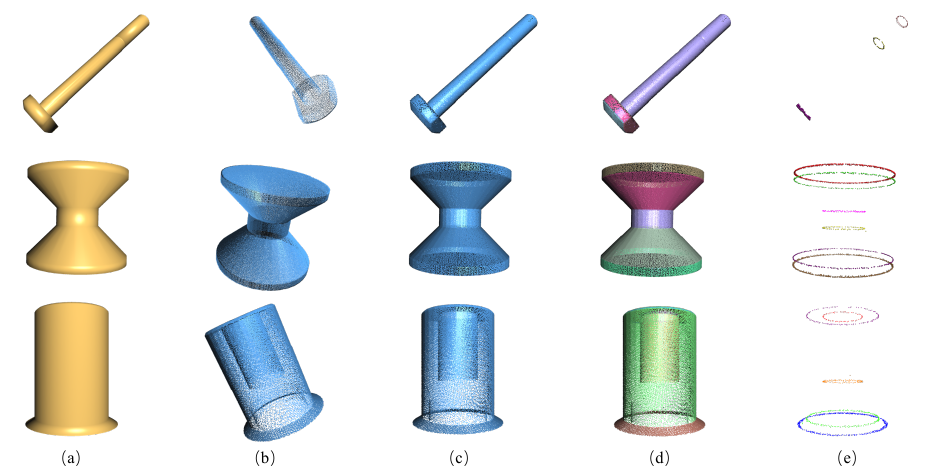}
	\caption{The error analysis results of our method.
  (a) Reference CAD model. (b) Scanned point cloud. (c) Registered point cloud. (d) Primitive detection result. (e) Multi-circle fitting result.}
 \label{fig:results_sup}
\end{figure*}

We have conducted error analysis on a variety of CAD models.
The visual results are shown in Figure~\ref{fig:results_sup}, and the manufacturing error is reported in Table~\ref{table:statistical_analysis_sup}.

We compute the registration error as the average distance between the registered point cloud and the sampling point cloud of the reference CAD model. 
As illustrated in Table~\ref{table:statistical_analysis_sup}, the registration error is an order of magnitude smaller than the manufacturing errors.
Based on the registered point clouds and the reference CAD mesh model, we conduct a hierarchical error analysis at three top-to-down levels: global, part, and feature.

\begin{table*}[htbp]
\centering
\caption{Manufacturing error (in millimeters) of aircraft CAD models.}
\label{table:statistical_analysis_sup}
\begin{tabular}{|c|c|c|c|c|c|c|c|c|}
  \hline
  & \multicolumn{1}{c|}{Registration error} & \multicolumn{1}{c|}{Global level} & \multicolumn{2}{c|}{Part level} & \multicolumn{4}{c|}{Feature level} \\ \cline{2-9}
  & $E_{\text{reg}}$ & $E_{\text{global}}$ & $E_{\text{part\_avg}}$ & $E_{\text{part\_max}}$ & $E_{\text{comp\_radius\_avg}}$ & $E_{\text{comp\_radius\_max}}$ & $E_{\text{comp\_centroid\_avg}}$ & $E_{\text{comp\_centroid\_max}}$ \\ \hline
(1) & 0.00824 & 0.0452 & 0.0561 & 0.0894 & 0.0678 & 0.0785 & 0.0513 & 0.0659 \\ \hline
(2) & 0.00765 & 0.0613 & 0.0724 & 0.0841 & 0.0626 & 0.0883 & 0.0631 & 0.0784 \\ \hline
(3) & 0.00437 & 0.0537 & 0.0612 & 0.0739 & 0.0657 & 0.0792 & 0.0593 & 0.0716 \\ \hline
\end{tabular}
\end{table*}

\end{document}